\documentclass{article}

\usepackage{iclr2027_conference,times}

\usepackage{cuted}
\usepackage{capt-of}

\usepackage{float}
\usepackage{graphicx}
\usepackage{url}
\usepackage{booktabs}
\usepackage{tabularx}
\usepackage{amsmath,amssymb,amsthm,bm}
\usepackage{microtype}
\usepackage[colorlinks=true,allcolors=blue]{hyperref}
\usepackage{algorithm}
\usepackage{algpseudocode}

\title{MA-UITD: Offline Multi-Agent Reward Inference for Characterizing Continuous Social Behavior}

\author{}

\author{
  Zeyuan Zhang\\
  Institute of Neuroscience, Chinese Academy of Sciences\\
  \texttt{zhangzeyuan@ion.ac.cn}
  \And
  Yiyun Yang\\
  Nankai University\\
  \texttt{1120210035@mail.nankai.edu.cn}
  \And
  Yuan Cheng\\
  Icahn School of Medicine at Mount Sinai\\
  \texttt{yuan.cheng@mssm.edu}
  \And
  Herbert Zheng Wu\\
  Icahn School of Medicine at Mount Sinai\\
  \texttt{zheng.wu@mssm.edu}
  \And
  Yusi Chen\\
  Institute of Neuroscience, Chinese Academy of Sciences\\
  \texttt{chenyusi@ion.ac.cn}
}

\iclrfinalcopy

\begin{document}

\maketitle
\lhead{}

\begin{abstract}
Multi-agent inverse reinforcement learning offers a powerful framework not only for imitation learning but also for understanding the neural computations supporting social interactions and decision-making. Inferring rewards from interacting agents is challenging because observed choices reflect both personal preferences and expectations about others. Existing methods often require additional environment interaction, repeated policy optimization as an inner loop, or restrictive assumptions about environment dynamics and action spaces, limiting their applicability to fixed behavioral datasets. We introduce Multi-Agent Unilateral Inverse Temporal Difference Learning (MA-UITD), an offline method for inferring explicit agent-specific rewards from joint demonstrations in continuous-state environments with either discrete or bounded continuous actions. Its core idea is to represent the long-term consequences of each agent’s own actions through belief-conditioned successor features that average over other agents’ behavior. Shared reward features and agent-specific preference vectors connect these unilateral values to the target reward function, while inverse temporal-difference consistency and noise-contrastive behavioral learning enable inference without known dynamics or an inner policy-optimization loop. Across simulated cooperative, competitive, and general-sum tasks, MA-UITD outperforms baselines in discrete-action settings and infers behaviorally compatible reward profiles in continuous-action settings. When applied to experimentally collected social-foraging trajectories, MA-UITD inferred distinct spatial and interaction-dependent reward structures for leaders and followers, providing a quantitative characterization of latent behavioral objectives that may be relevant for understanding the neural computations underlying social behavior. We make our code available at \url{https://github.com/ZaynZh/MA-UITD}
\end{abstract}

\section{Introduction}

A central goal of behavioral science and neuroscience is to understand the objectives that organize individual behavior. Although accurate action prediction is an important benchmark, an \textit{explicit reward} function provides a more interpretable computational account by specifying which outcomes an animal values, how it trades off competing outcomes, and how these valuations change with experience. Subjective-value signals have been identified in orbitofrontal cortex and across a broader valuation network involving ventromedial prefrontal cortex and ventral striatum \citep{padoa_schioppa_neurons_2006,bartra_valuation_2013}, whereas phasic activity of midbrain dopamine neurons has been linked to quantitative reward-prediction errors and corresponding distribution \citep{schultz_neural_1997,bayer_midbrain_2005,dabney_distributional_2020}. Consequently, inferred rewards—and the values and prediction errors they imply—can serve as trial-by-trial computational variables for model-based analyses of neural recordings, enabling tests of latent-computation hypotheses beyond action prediction alone \citep{odoherty_model-based_2007}.

Inverse reinforcement learning (IRL) provides a principled framework for recovering such latent objectives by inferring reward functions under which observed behavior is likely. In single-agent settings, IRL has been used to recover both stable and dynamically evolving goals from behavior \citep{ziebart2008maximum,ashwood2022dynamic,wu2025distributionalirl,ke2025swirl, wang_learning_2025}. Model-derived reward and value representations have also begun to provide computational variables for characterizing neural activity and examining how valuation-related computations may be altered in neuropsychiatric disorders \citep{bromberg-martin_neural_2024,moneta_representational_2024}. However, conventional IRL was developed primarily for single-agent settings and treated each individual as an isolated decision-maker, making it difficult to distinguish intrinsic preferences from socially contingent choices that depend on other individuals. 

Inferring these objectives becomes substantially more challenging when individuals interact. In social settings, an animal’s choice depends on both its own preferences and its expectations about others. Approaching a shared resource, for example, may reflect one's own resource preference and anticipated interaction with other agents such as cooperation and competition. Similar observed actions can consequently arise from different incentives. Computational models have framed such social inference as IRL in a partially observable Markov decision process (POMDP), where a focal agent acts according to beliefs about an incompletely observed environment \citep{jara2019theory, baker_rational_2017, velez2017interpreting}. Although this formulation captures uncertainty and false beliefs, other individuals are typically incorporated into the focal agent’s environment rather than represented as autonomous decision-makers with their own actions, policies, and rewards. By contrast, a multi-agent MDP, or Markov game, explicitly represents multiple decision-makers whose joint actions determine state transitions and whose policies are mutually interdependent. Multi-agent inverse reinforcement learning (MAIRL) builds on this formulation to infer agent-specific rewards while accounting for how each agent anticipates and responds to the behavior of others \citep{yu_multi-agent_2019,tian_learning_2021,neumeyer_general-sum_2021,Chen2024.10.09.617461}. MAIRL thus offers a systematic framework for separating what each individual values from what it expects others to do, providing interpretable computational variables for studying cooperation, competition, and general-sum interactions.

Applying MAIRL to animal behavior imposes several practical constraints: experiments typically yield a fixed dataset of joint trajectories without reward labels, known transition dynamics, or opportunities to collect additional behavior under a learned policy, and the recorded states and actions are often continuous. Existing methods address parts of this setting: Multi-Agent Adversarial Inverse Reinforcement Learning(MA-AIRL) combines reward estimation with adversarial policy optimization, requiring repeated policy updates and environment interaction during training \citep{yu_multi-agent_2019}. Multi-Agent Marginal Q-Learning(MAMQL) avoids adversarial training by learning through opponent-marginalized values, using discrete actions and additional policy rollouts in its reported training procedure \citep{haynam_multi-agent_2025}. These constraints motivate a fully offline approach that infers explicit agent-specific rewards directly from fixed joint demonstrations, without known dynamics or an inner policy-optimization loop, while supporting \textit{continuous behavior}.

We introduce Multi-Agent Unilateral Inverse Temporal Difference Learning (MA-UITD), an offline method for inferring explicit agent-specific rewards from joint demonstrations. We focus on arrival-state rewards that describe the value individuals assign to attained environmental and social states. The inferred rewards provide interpretable explanations of behavior while supporting expert-level behavioral imitation. Building on successor features and inverse temporal-difference learning \citep{filos_psiphi-learning_2021}, our main contributions are:

\begin{itemize}
    \item We introduce belief-conditioned successor features (BCSF), which marginalize over other agents' anticipated behavior to represent unilateral long-term values and recover interpretable agent-specific rewards.
    \item We develop an offline learning procedure that combines inverse temporal-difference consistency with noise-contrastive behavioral learning, requires no known dynamics or policy-optimization loop, and supports \textit{continuous states} with either discrete or \textit{continuous actions}.
    \item Across cooperative, competitive, and general-sum simulations, MA-UITD outperforms discrete-action baselines and captures task-relevant reward structure in continuous-action settings.
    \item Applied to animal social-interaction data, MA-UITD reveals distinct spatial and interaction-dependent incentives underlying leader--follower dynamics.
\end{itemize}

\vspace{-1em}
\section{Related Work}
\paragraph{Multi-agent inverse reward learning.}
In multi-agent settings, MA-AIRL \citep{yu_multi-agent_2019} couples agent-specific reward estimation to an adversarial discriminator and policy optimization. Such adversarial approaches involve an inner policy loop and outer loop, which could be computationally expensive and unstable. MAMQL \citep{haynam_multi-agent_2025} instead learns opponent-marginalized reward through a non-adversarial inverse soft-$Q$ objective; however, it assumes a discrete action set and obtains explicit rewards through a separate reconstruction loss, with its reported training procedure drawing additional policy rollouts. Other work models strategic beliefs through latent quantal level-$k$ reasoning \citep{tian_learning_2021}, but requires recursive value iteration and identifies continuous state and action spaces as a limitation. GS-CIOC \citep{neumeyer_general-sum_2021} supports non-adversarial reward inference with continuous states and actions, yet presumes known dynamics and relies on local linear-quadratic approximations around demonstrations under a specified equilibrium model.

\paragraph{Successor-feature-based inverse reward learning.}
A complementary approach connects explicit rewards to long-term values through a shared feature representation. \citet{filos_psiphi-learning_2021} introduce inverse temporal-difference (ITD) learning, which jointly estimates shared cumulants, demonstrator-specific successor features, and preference vectors from fixed demonstrations without reward labels. By combining behavioral likelihood fitting with successor-feature Bellman consistency, ITD yields explicit rewards from the learned cumulants and preferences. However, its formal setup considers different reward tasks within a common controlled Markov process with discrete actions, without an explicit Markov-game formulation of strategic interactions.


\paragraph{Offline reward inference for animal behavior.}
These considerations are particularly important for studies of animal behavior, where joint trajectories can be recorded experimentally but constructing a faithful interactive simulator is often difficult because the underlying behavioral dynamics are only partially known \citep{fujii_data-driven_2025}. In such settings, methods that require repeated environment interaction, additional policy rollouts, or an accurate model of the dynamics may therefore be impractical. Offline reward-inference methods can recover explicit rewards from fixed demonstrations without known dynamics or additional environment interaction, but are formulated for single-agent settings with discrete actions \citep{chan_bayesianIRL_2021}. This motivates methods that can infer explicit, agent-specific rewards directly from fixed datasets of continuous joint trajectories in multi-agent systems.

Building on these, our method learns belief-conditioned successor features directly from a fixed dataset of joint transitions and uses noise-contrastive estimation to avoid the intractable continuous-action normalizer. It thereby recovers an explicit agent-specific reward profile in continuous state and action spaces, without adversarial optimization, additional environment interaction, or a forward policy-optimization loop; moreover, equilibrium and finite-depth reasoning enter as restrictions on a common predictive belief rather than as hard-coded solution concepts.

\section{Problem Formulation}

\paragraph{Markov game and offline demonstrations.}
We consider an $n$-agent discounted Markov game without known reward
functions
\(\mathcal{G}^{-R}=\left(\mathcal{N},\mathcal{S},\{\mathcal{A}_i\}_{i=1}^{n},P,p_0,\gamma\right),\)
where $\mathcal{N}=\{1,\ldots,n\}$ is the set of agents,
$\mathcal{S}$ is the joint state space,
$\mathcal{A}_i$ is the action space of agent $i$,
$\mathcal{A}=\prod_{i=1}^{n}\mathcal{A}_i$ is the joint action space,
$P(s' \mid s,\boldsymbol{a})$ denotes the transition dynamics,
$p_0$ is the initial-state distribution, and
$\gamma\in[0,1)$ is the discount factor.
We allow each $\mathcal{A}_i$ to be either discrete or a bounded
continuous subset of $\mathbb{R}^{d_i}$.

At time $t$, all agents simultaneously choose a joint action
$\boldsymbol{a}_t=(a_{1,t},\ldots,a_{n,t})$, after which the environment
transitions according to $s_{t+1}\sim P(\cdot\mid s_t,\boldsymbol{a}_t)$. We are given only an offline dataset of joint expert trajectories
\[
\mathcal{D}
=
\left\{
\tau^{(m)}
\right\}_{m=1}^{M},
\qquad
\tau^{(m)}
=
\left(
s_0^{(m)}, \boldsymbol{a}_0^{(m)},
s_1^{(m)}, \boldsymbol{a}_1^{(m)}
\ldots,
s_{T_m}^{(m)}, \boldsymbol{a}_{T_m}^{(m)}
\right),
\]
without observing the rewards that generated the demonstrated behavior.
No additional interaction with the environment is assumed during
inverse learning.

\paragraph{Arrival-state rewards.}
Motivated by behavioral and neuroscience settings in which the inferred reward is interpreted as the value assigned to attained outcomes or configurations, we focus on agent-specific arrival-state rewards. That is, the reward received by agent $i$ following a transition $(s_t,\mathbf{a}_t,s_{t+1})$ is assumed to take the form
\[
R_i(s_t,\boldsymbol{a}_t,s_{t+1})
=
r_i(s_{t+1}),
\qquad
r_i:\mathcal{S}\rightarrow\mathbb{R}.
\]
Because $s$ is the joint environmental state, $r_i(s)$ may still encode
both individual preferences and interactions among agents. Thus, this
assumption restricts where reward is assigned---to the resulting state
rather than directly to the executed action---without requiring agents
to share the same reward function. When deterministic choices produce distinguishable outcome
states, action-contingent rewards can be represented equivalently as
arrival-state rewards. This includes common cognitive tasks such as
multi-armed bandits and the prisoner's dilemma when the outcome state records
the selected action or joint action.


\paragraph{Reward compatibility rather than reward identification.}
Inverse reinforcement learning is generally non-identifiable, and this ambiguity is further compounded in multi-agent settings. In particular, rewards in general-sum Markov games are not uniquely identifiable from demonstrations without additional assumptions \citep{freihaut_feasible_2025}. We therefore seek a representative agent-specific reward profile
\(\widehat{R}=(\hat r_1,\ldots,\hat r_n)\) within the arrival-state reward class, rather than claim recovery of the unique ground truth. We assess compatibility operationally: under agent \(i\)'s predictive belief \(\boldsymbol\mu_{i,-i}\), the reward-induced belief-conditioned score should assign higher relative likelihood to demonstrated actions than to counterfactual alternatives. Equilibrium and finite-depth reasoning enter only through the choice of this belief, rather than as imposed solution concepts.

\section{Multi-Agent Unilateral Inverse Temporal Difference Learning
(MA-UITD)}

\label{sec:method}
We introduce MA-UITD, an offline method for inferring agent-specific rewards from fixed joint demonstrations. It involves three key components: a shared reward feature map $\Phi$, an agent-specific preference vector $w_i$, and agent-specific belief-conditioned successor features (BCSFs). BCSFs represent the long-term
consequences of an agent's own action after averaging over its prediction of the other agents' behavior. This separates strategic assumptions, encoded by the predictive belief, from reward preferences, encoded by $w_i$.

Learning combines two complementary objectives. A contrastive behavioral
loss makes the inferred values assign greater relative likelihood to
demonstrated actions than to sampled alternatives, while a unilateral inverse
temporal-difference loss enforces Bellman consistency between the BCSFs and
the learned reward features. Both objectives are estimated from the fixed
dataset, without a transition model, additional environment interaction, or
an inner policy-optimization loop. We first define the BCSF representation
and then present the offline learning procedure.

\subsection{Belief-Conditioned Successor Features (BCSF)}
\label{sec:bcsf}

For a stationary joint policy $\boldsymbol{\pi}$, define agent $i$'s
joint-action value as
\begin{equation}
Q_i^{\boldsymbol{\pi}}(s,\boldsymbol a)
\triangleq
\mathbb{E}_{\boldsymbol{\pi},P}
\left[
\left.
\sum_{t=0}^{\infty}\gamma^t r_i(s_{t+1})
\,\right|\,
s_0=s,\boldsymbol a_0=\boldsymbol a
\right].
\label{eq:joint_q}
\end{equation}
This value conditions on the complete joint action
$\boldsymbol a=(a_i,\boldsymbol a_{-i})$. The initial joint
action is fixed, while subsequent joint actions are generated by
$\boldsymbol{\pi}$ and states evolve according to $P$. In our setting, the demonstrations consist of trajectories generated by an
unknown stationary joint policy $\boldsymbol{\pi}^{D}$. 

The joint-action value
$Q_i^{\boldsymbol{\pi}}(s,a_i,\boldsymbol a_{-i})$ evaluates a fully
specified joint action. In a simultaneous-move interaction, however,
agent $i$ must select $a_i$ without observing the other agents'
current actions. Its policy therefore cannot respond directly to the
realized $\boldsymbol a_{-i}$. Instead, evaluating $a_i$ requires a
belief about the actions of the other agents. Let
$\boldsymbol{\mu}_{i,-i}(\boldsymbol a_{-i}\mid s)$ denote its belief about the
other agents' actions, and let
$\boldsymbol{\pi}^{(i)}=(\pi_i,\boldsymbol{\mu}_{i,-i})$ denote its
perceived continuation profile. The value relevant to its choice is
therefore
\begin{equation}
\bar Q_i^{\pi_i,\boldsymbol{\mu}_{i,-i}}(s,a_i)
\triangleq
\mathbb{E}_{
\boldsymbol a_{-i}\sim
\boldsymbol{\mu}_{i,-i}(\cdot\mid s)}
\left[
Q_i^{\boldsymbol{\pi}^{(i)}}
    (s,a_i,\boldsymbol a_{-i})
\right].
\label{eq:belief_q}
\end{equation}

We assume an arrival-state linear reward representation,
\begin{equation}
r_i(s_{t+1})=\Phi(s_{t+1})^\top w_i.
\label{eq:linear_reward}
\end{equation}
The corresponding joint-action successor features are
\begin{equation}
\Psi_i^{\boldsymbol{\pi}^{(i)}}(s,\boldsymbol a)
\triangleq
\mathbb{E}_{\boldsymbol{\pi}^{(i)},P}
\left[
\left.
\sum_{t=0}^{\infty}\gamma^t\Phi(s_{t+1})
\,\right|\,
s_0=s,\boldsymbol a_0=\boldsymbol a
\right],
\label{eq:joint_sf}
\end{equation}
so that
$Q_i^{\boldsymbol{\pi}^{(i)}}(s,\boldsymbol a)
=
\Psi_i^{\boldsymbol{\pi}^{(i)}}(s,\boldsymbol a)^\top w_i$.
Marginalizing these successor features over the agent's belief gives
the \emph{belief-conditioned successor features} (BCSFs):
\begin{equation}
\bar\Psi_i^{\pi_i,\boldsymbol{\mu}_{i,-i}}(s,a_i)
\triangleq
\mathbb{E}_{
\boldsymbol a_{-i}\sim
\boldsymbol{\mu}_{i,-i}(\cdot\mid s)}
\left[
\Psi_i^{\boldsymbol{\pi}^{(i)}}
    (s,a_i,\boldsymbol a_{-i})
\right].
\label{eq:bcsf}
\end{equation}
By linearity,
\begin{equation}
\bar Q_i^{\pi_i,\boldsymbol{\mu}_{i,-i}}(s,a_i)
=
\bar\Psi_i^{\pi_i,\boldsymbol{\mu}_{i,-i}}(s,a_i)^\top w_i.
\label{eq:bcsf_q}
\end{equation}

We model agent $i$'s policy as a logit response to this unilateral
value:
\begin{equation}
\pi_i(a_i\mid s)
=
\frac{
\exp\left(
\lambda_i
\bar\Psi_i^{\pi_i,\boldsymbol{\mu}_{i,-i}}(s,a_i)^\top w_i
\right)
}{
\displaystyle
\int_{\mathcal A_i}
\exp\left(
\lambda_i
\bar\Psi_i^{\pi_i,\boldsymbol{\mu}_{i,-i}}(s,\tilde a_i)^\top w_i
\right)
\,d\tilde a_i
}.
\label{eq:bcsf_policy}
\end{equation}
For discrete action spaces, the integral is replaced by a sum over
$\mathcal A_i$.

The BCSFs satisfy the Bellman recursion
\begin{align}
\bar\Psi_i^{\pi_i,\boldsymbol\mu_{i,-i}}(s,a_i)
={}&
\mathbb{E}_{\substack{
\boldsymbol a_{-i}\sim\boldsymbol\mu_{i,-i}(\cdot\mid s),\\
s'\sim P(\cdot\mid s,a_i,\boldsymbol a_{-i}),\\
a_i'\sim\pi_i(\cdot\mid s')
}}
\left[
\Phi(s')
+
\gamma
\bar\Psi_i^{\pi_i,\boldsymbol\mu_{i,-i}}(s',a_i')
\right].
\label{eq:bcsf_bellman}
\end{align}
This recursion integrates over the other agents' current actions, the next
state, and agent $i$'s continuation action. It is the basis of the offline
temporal-difference objective introduced below.

The belief $\boldsymbol\mu_{i,-i}$ determines the strategic interpretation
of the representation. For a logit quantal-response equilibrium (LQRE)
\citep{mckelvey_quantal_1995}, there is a common policy profile
$\boldsymbol\pi^{\star}$ such that, for every agent,
\begin{equation}
\boldsymbol{\mu}_{i,-i}
=
\boldsymbol{\pi}_{-i}^{\star}
=
\prod_{j\neq i}\pi_j^{\star},
\qquad
\pi_i=\pi_i^{\star}.
\label{eq:lqre_specialization}
\end{equation}
All perceived continuation profiles then coincide with
$\boldsymbol\pi^{\star}$, and Equation~\eqref{eq:bcsf_policy} holds
simultaneously for every agent. Under quantal level-$k$ reasoning, the same
representation instead uses
$\boldsymbol\mu_{i,-i}^{(k)}=\boldsymbol\pi_{-i}^{(k-1)}$. Thus, equilibrium
and finite-depth reasoning correspond to different restrictions on the
predictive belief rather than different reward representations.

\subsection{Offline Reward Inference with MA-UITD}
\label{sec:offline_ma-uitd}
Given demonstrations $\mathcal D$, we parameterize the shared reward feature
map as $\Phi(s;\boldsymbol\theta^\Phi)$, agent $i$'s BCSFs as
$\bar\Psi_i(s,a_i;\boldsymbol\eta_i)$, and its preference vector as $w_i$.
For offline estimation, we take the predictive belief to be calibrated to
the demonstrated conditional behavior of the other agents $\boldsymbol\mu_{i,-i}(\boldsymbol a_{-i}\mid s)
\approx
\boldsymbol\pi^D_{-i}(\boldsymbol a_{-i}\mid s)$. Observed joint transitions then provide Monte Carlo samples of the
expectations in Equation~\eqref{eq:bcsf_bellman}. We fit the parameterized
representation using a contrastive behavioral objective and a BCSF
temporal-difference objective.

\subsubsection{Contrastive behavioral learning}
The normalizing integral in Equation~\eqref{eq:bcsf_policy} is generally
intractable for continuous actions. We therefore fit the unnormalized BCSF
scores using noise-contrastive estimation (NCE)
\citep{gutmann2010noise}. At a demonstrated state $s$, let $a_i^+$ denote
agent $i$'s observed action and sample $K$ alternative actions
$a_{i,1:K}^-\sim\nu_i(\cdot\mid s)$ from a declared proposal distribution
over $\mathcal A_i$. Their scores are $q_i^+=\bar\Psi_i(s,a_i^+;\boldsymbol\eta_i)^\top w_i, q_{i,k}^-=\bar\Psi_i(s,a_{i,k}^-;\boldsymbol\eta_i)^\top w_i.$

We minimize the contrastive loss
\begin{equation}
\begin{aligned}
    \mathcal{L}_i^{\mathrm{NCE}}(\boldsymbol\eta_i,w_i)
    \triangleq
    -\mathbb{E}_{\substack{
        (s,a_i^{+})\sim\mathcal{D},\\
        a_{i,1:K}^{-}\sim\nu_i(\cdot\mid s)
    }}
    \left[
        \log
        \frac{
            \exp(q_i^{+}/\tau_{\mathrm{NCE}})
        }{
            \exp(q_i^{+}/\tau_{\mathrm{NCE}})
            +
            \sum_{k=1}^{K}
            \exp(q_{i,k}^{-}/\tau_{\mathrm{NCE}})
        }
    \right].
    \label{eq:bc-nce-loss}
\end{aligned}
\end{equation}
where $\tau_{\mathrm{NCE}}>0$ controls the assumed stochasticity of the
demonstrated behavior and absorbs the rationality scale $\lambda_i$ in
Equation~\eqref{eq:bcsf_policy}. A lower temperature emphasizes the
highest-valued candidate, whereas a higher temperature represents noisier
action selection.
This objective compares only unilateral alternatives for agent $i$; the
other agents' behavior remains marginalized through $\bar\Psi_i$.

\subsubsection{Unilateral inverse temporal-difference learning}
Inverse temporal-difference learning connects reward features to successor
features through their Bellman recursion
\citep{filos_psiphi-learning_2021}. In our setting, the recursion must be
enforced on BCSFs because these are the unilateral quantities that determine
the behavioral scores in Equation~\eqref{eq:bc-nce-loss}.

Consider consecutive demonstrated transitions
$(s_t,\boldsymbol a_t,s_{t+1},\boldsymbol a_{t+1})\sim\mathcal D$. Under
Section~\ref{sec:offline_ma-uitd}, the observed
$\boldsymbol a_{-i,t}$ samples the current belief, $s_{t+1}$ samples the
unknown transition kernel, and $a_{i,t+1}$ samples agent $i$'s continuation
policy. The one-step target
\begin{equation}
y_{i,t}
=
\Phi(s_{t+1};\boldsymbol\theta^\Phi)
+
\gamma
\operatorname{stopgrad}\!\left(
\bar\Psi_i(s_{t+1},a_{i,t+1};\tilde{\boldsymbol\eta}_i)
\right)
\label{eq:uitd_target}
\end{equation}
is therefore a Monte Carlo estimate of the right-hand side of
Equation~\eqref{eq:bcsf_bellman}. Here
$\tilde{\boldsymbol\eta}_i$ denotes target-network parameters, which are
treated as fixed when differentiating the loss. We minimize
\begin{align}
\mathcal{L}^{\mathrm{UITD}}_i
(\boldsymbol{\theta}^{\Phi},\boldsymbol{\eta}_i)
\triangleq
&\mathbb{E}_{
(s_t,\boldsymbol{a}_t,s_{t+1},\boldsymbol{a}_{t+1})
\sim\mathcal{D}
}
\Bigg[
\Bigg\|
\bar\Psi_i(s_t,a_{i,t};\boldsymbol\eta_i)
-
y_{i,t}
\Bigg\|^2
\Bigg].
\label{eq:UITD_loss}
\end{align}
This sample-based objective implicitly averages over the other agents'
actions and the environment transition, so neither the transition kernel nor
an explicit opponent model must be estimated.

\subsubsection{Optimization and reward recovery}
Training alternates between the two objectives. See Algorithm~\ref{alg:ma-uitd} for a detailed implementation. The NCE step updates
$\boldsymbol\eta_i$ and $w_i$ to fit the demonstrated action preferences,
with an $\ell_1$ penalty $\lambda_w\lVert w_i\rVert_1$. The UITD step updates
$\boldsymbol\theta^\Phi$ and every $\boldsymbol\eta_i$ to enforce temporal
consistency, followed by an update of the target parameters
$\tilde{\boldsymbol\eta}_i$. After convergence, the agent-specific rewards
are recovered directly as $\widehat r_i(s)
=
\Phi(s;\boldsymbol\theta^\Phi)^\top w_i$.
Because each agent has its own $w_i$, the same representation applies to
cooperative, competitive, and general-sum games.

\section{Results}
We evaluate MA-UITD in three simulated multi-agent environments spanning cooperative, competitive, and general-sum interactions. Our experiments examine 1) whether MA-UITD reproduces expert-level behavior and 2) recovers interpretable rewards compatible with the demonstrations. We first compare MA-UITD with baselines in discrete-action settings and then evaluate it in continuous-action settings.

\subsection{Experimental setup}

\textbf{Environments.} We evaluate three games: (1) cooperative \textit{Simple Spread} from MPE2 \citep{lowe2017mpe}, where two agents cooperate to cover two landmarks; (2) competitive \textit{Simple Push}, where a good agent attempts to reach a target landmark while an adversary tries to prevent it from doing so through physical interaction; and (3) general-sum \textit{Gems}, adapted from \citet{haynam_multi-agent_2025}. In \textit{Gems}, Red and Blue maximize individual returns by collecting their respective private gems for \(1\) reward or jointly collecting a purple gem for \(6\), after which the episode terminates. Agents choose among nine directional actions in the discrete setting or bounded two-dimensional displacements in the continuous setting, with an \(80\)-step horizon.

\textbf{Baselines.} We evaluate MA-UITD against two baselines. The first is MA-AIRL \citep{yu_multi-agent_2019}, an adversarial learning-based multi-agent IRL method. Then, we benchmark against MAMQL \citep{haynam_multi-agent_2025}, an IQ-Learn-style MAIRL algorithm that similarly marginalizes the joint-action value function over the opponents’ continuation policies but is designed only for discrete action spaces.

\subsection{Policy and reward recovery}
\paragraph{Policy imitation.} MA-UITD achieved higher recovered-policy returns than both baselines in \textit{Simple Spread} and for both roles in \textit{Simple Push}, while most closely approaching expert performance in \textit{Gems} (Table~\ref{tab:recovered-policy-return-4000}). In \textit{Gems}, MA-UITD and MAMQL recovered both cooperative and competitive behaviors, whereas MA-AIRL converged exclusively to competitive behavior (Table~\ref{tab:Gems_collection_distribution}).

\vspace{-1em}

\begin{table*}[htbp]
  \centering
  \caption{Policy imitation performance measured as mean true return on discrete-action tasks}
  \label{tab:recovered-policy-return-4000}
  \small
  \setlength{\tabcolsep}{7pt}
  \begin{tabular}{@{}lcccc@{}}
    \toprule
    Method
      & Simple Spread
      & \multicolumn{2}{c}{Simple Push cross-play}
      & Gems \\
    \cmidrule(lr){3-4}
      & Shared return $\uparrow$
      & Learned adversary $\uparrow$
      & Learned good $\uparrow$
      & Agent-mean return $\uparrow$\\
    \midrule
    Expert reference
      & $-21.592$
      & $0.722$
      & $-7.228$
      & $3.581$ \\
    MA-AIRL
      & $-22.135 \pm 0.071$
      & $-0.041 \pm 0.005$
      & $-7.885 \pm 0.081$
      & $0.516 \pm 0.004$ \\
    MAMQL
      & $-28.098 \pm 0.415$
      & $0.468 \pm 0.003$
      & $-7.443 \pm 0.008$
      & $2.587 \pm 0.542$ \\
    MA-UITD
      & $\mathbf{-21.848 \pm 0.033}$
      & $\mathbf{0.586 \pm 0.021}$
      & $\mathbf{-7.319 \pm 0.022}$
      & $\mathbf{2.953 \pm 0.099}$ \\
    \bottomrule
  \end{tabular}
  \vspace{2pt}
  {\footnotesize Values are reported as mean $\pm$ sample standard deviation across three training seeds, each evaluated over 500 stochastic-policy episodes.}
\end{table*}

\vspace{-2em}


\vspace{1em}

\begin{figure}[htbp]
    \centering
    \includegraphics[width=\linewidth]{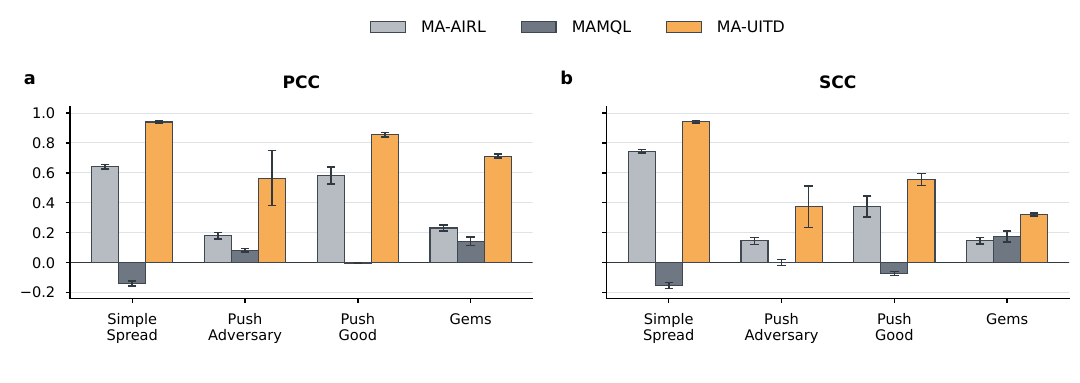}
    \par\kern-15pt
    \caption{\textbf{Reward recovery for MA-UITD and baselines.} Each method is independently trained with three random seeds, and correlations are computed on 500 held-out test episodes for each trained model.}
    \label{fig:reward_recovery_discrete}
\end{figure}

\paragraph{Reward recovery.} 
We evaluate reward recovery on held-out transitions using Pearson’s correlation coefficient (PCC), which measures linear agreement, and Spearman’s rank correlation coefficient (SCC), which measures agreement in reward ordering. For each agent, we compare the ground-truth immediate reward with the reward predicted by the learned reward function (Table~\ref{tab:PCC/SCC_comparison}). Figure~\ref{fig:reward_recovery_discrete} shows that MA-UITD outperforms the baselines on both measures across all three simulated environments.

\begin{figure}[htbp]
    \centering
    \includegraphics[width=\linewidth]{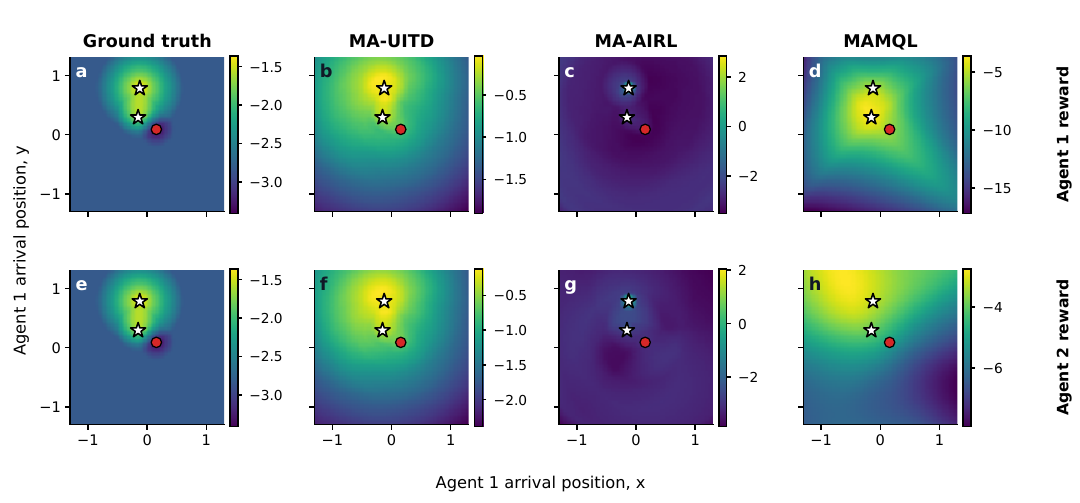}
    \par\kern-15pt
    \caption{\textbf{Visualization of the ground-truth and recovered reward landscapes in \textit{Simple Spread}}. Agent 1's arrival position is varied over the spatial grid while agent 2 (red circle), the landmarks (stars), and arrival velocities are held fixed. The top and bottom rows show the rewards recovered for agents 1 and 2, respectively.}
    \label{fig:simple_spread_visualization}
\end{figure}

\textbf{Interpretable reward visualization.} We next visualize the reward functions recovered by each method. In \textit{Simple Spread} (Fig.~\ref{fig:simple_spread_visualization}), MA-UITD captures the task's underlying coverage structure: once one agent is close to a landmark, the other landmark becomes relatively more rewarding for the other agent, encouraging the agents to distribute across landmarks rather than collapse onto the same target. Consistent with the task’s symmetric cooperative objective, MA-UITD recovers nearly identical spatial reward structures for both agents. By contrast, MA-AIRL produces highly localized peaks around individual landmarks, whereas MAMQL assigns high rewards to regions inconsistent with the ground-truth structure and recovers markedly different reward functions across agents. Additional visualizations for \textit{Simple Spread} and the other tasks are provided in Appendix~\ref{appendix_discrete_reward_visualizations}.

\paragraph{Continuous action spaces.} We next evaluated MA-UITD in \textit{continuous-action} versions of all three environments. The recovered policies achieved returns matching the expert reference in \textit{Gems} and approached expert performance in \textit{Simple Spread}, whereas performance in the competitive \textit{Simple Push} task remained role dependent (Table~\ref{tab:ma-uitd-continuous-policy}). On held-out trajectories, the recovered rewards showed consistent positive agreement with the ground-truth rewards across all three tasks, with the strongest linear agreement observed in \textit{Simple Spread} (agent-mean PCC $0.828 \pm 0.033$; Table~\ref{tab:ma-uitd-continuous-reward}). The comparatively modest SCC values likely reflect the prevalence of transitions with similar immediate rewards in continuous environments, where small prediction errors can reverse local rankings. We also conducted a qualitative visual analysis by comparing the recovered and ground-truth reward landscapes (Appendix~\ref{appendix_continuous_reward_visualizations}). Together, these quantitative and visual results provide evidence that MA-UITD can infer behaviorally compatible reward profiles directly from offline demonstrations with bounded continuous actions.

\begin{table*}[htbp]
  \centering
  \caption{Policy imitation performance of MA-UITD on continuous-control tasks.}
  \label{tab:ma-uitd-continuous-policy}
  \small
  \setlength{\tabcolsep}{7pt}
  \begin{tabular}{@{}lcccc@{}}
    \toprule
    Method
      & Simple Spread
      & \multicolumn{2}{c}{Simple Push cross-play}
      & \textit{Gems} \\
    \cmidrule(lr){3-4}
      & Shared return $\uparrow$
      & Learned adversary $\uparrow$
      & Learned good $\uparrow$
      & Agent-mean return $\uparrow$ \\
    \midrule
    Expert reference
      & $-33.338$
      & $-0.063$
      & $-6.707$
      & $3.337$ \\
    MA-UITD
      & $-36.783 \pm 0.100$
      & $-2.286 \pm 0.179$
      & $-8.611 \pm 0.396$
      & $3.983 \pm 0.026$ \\
    \bottomrule
  \end{tabular}
  \vspace{2pt}

  {\footnotesize Values are reported as mean $\pm$ sample standard deviation across three training seeds, each evaluated over 500 stochastic-policy episodes. }
\end{table*}

\begin{table*}[htbp]
  \centering
  \caption{Reward recovery of MA-UITD on continuous-control tasks with held-out demonstrations.}
  \label{tab:ma-uitd-continuous-reward}
  \small
  \setlength{\tabcolsep}{6pt}
  \begin{tabular}{@{}lccccc@{}}
    \toprule
    Metric
      & Simple Spread
      & \multicolumn{2}{c}{Simple Push}
      & \multicolumn{2}{c}{\textit{Gems}} \\
    \cmidrule(lr){3-4}\cmidrule(lr){5-6}
      & Agent mean
      & Adversary
      & Good
      & Red
      & Blue \\
    \midrule
    PCC $\uparrow$
      & $0.828 \pm 0.033$
      & $0.339 \pm 0.022$
      & $0.717 \pm 0.219$
      & $0.663 \pm 0.129$
      & $0.643 \pm 0.139$ \\
    SCC $\uparrow$
      & $0.388 \pm 0.089$
      & $0.322 \pm 0.093$
      & $0.396 \pm 0.035$
      & $0.294 \pm 0.028$
      & $0.305 \pm 0.010$ \\
    \bottomrule
  \end{tabular}
  \vspace{2pt}

  {\footnotesize Correlations are computed over all transitions in each seed’s held-out 500-episode test split and reported as mean $\pm$ sample standard deviation across three seeds.}
\end{table*}

\begin{figure}[htbp]
    \centering
    \includegraphics[width=\linewidth]{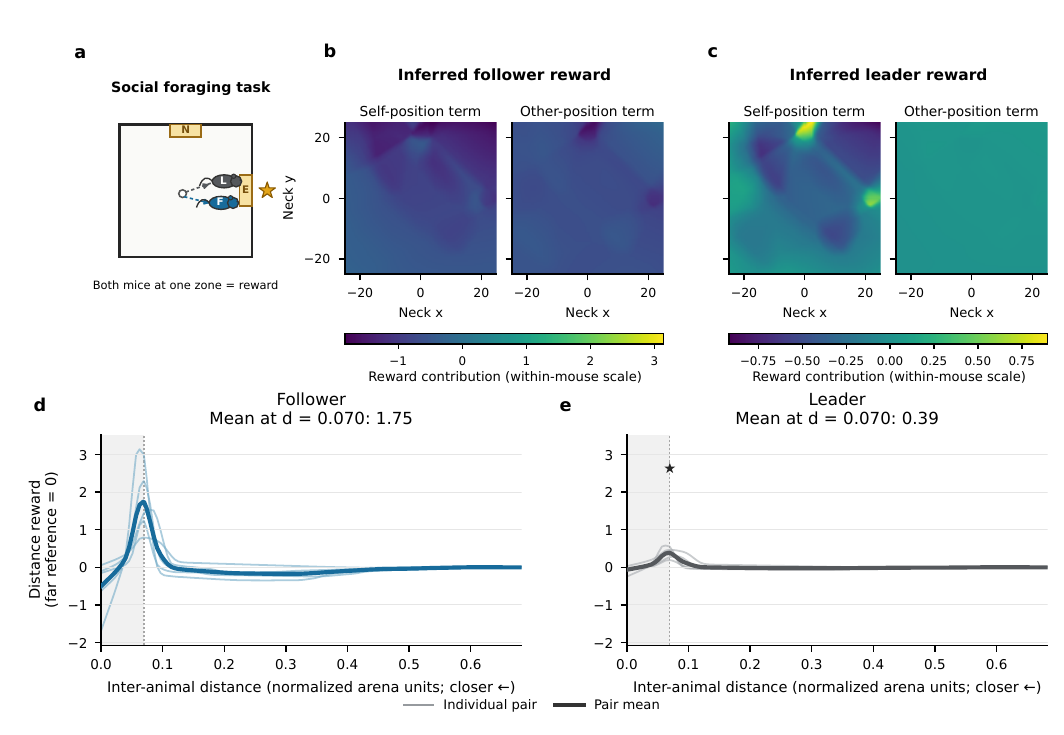}
    \par\kern-15pt
    \caption{\textbf{Learned position and proximity reward components.}
    \textbf{a}, Social-foraging task.
    \textbf{b,c}, Example self- and other-position reward maps, scaled with the distance component using a common within-animal reward range.
    \textbf{d,e}, Inferred distance-dependent rewards for individual pairs (thin) and their mean (thick). Gray shading marks the range below the common 5th-percentile threshold, where data were sparsely sampled. Followers valued proximal partner locations more highly than leaders (two-sided paired $t$-test at $d=0.07$, $P=0.026$; $n=5$ pairs).}
    \label{fig:leader_follower}
\end{figure}

\paragraph{Application to real animal trajectories without trajectory binning.}
We next evaluated MA-UITD using joint trajectories from pairs of mice performing a social-foraging task \citep{ChengAsymmetric}. In this task, the two animals had to cooperate by navigating to the same reward zone to obtain reward (Fig.~\ref{fig:leader_follower}a).  Over training, pairs developed stable asymmetric behavioral roles, with one animal acting as a leader and the other as a follower, potentially reflecting distinct internal reward functions shaped by uncertainty under decentralized, simultaneous movement. Because MA-UITD models a separate reward function for each agent, it can directly represent this form of behavioral asymmetry. Without prior knowledge, however, an animal's internal reward could depend on arbitrary task variables, leaving inverse reward inference severely underconstrained. We therefore restricted the reward representation to the two animals' positions in the arena and their inter-animal distance, defining reward functions on separate self-position, other-position, and distance components.

Figure~\ref{fig:leader_follower}b,c shows distinct learned reward structures for a representative leader--follower pair. The leader's learned reward depended primarily on its own spatial location, whereas the follower's position-dependent contributions were weaker relative to its proximity-dependent component. This pattern was consistent across mouse pairs: follower rewards increased strongly at short inter-animal distances, while leader rewards showed substantially weaker modulation by distance (Fig.~\ref{fig:leader_follower}d,e). These observations support leader--follower dynamics in which the leader navigates according to spatial goals and the follower is motivated to remain close to the leader.

\section{Discussion}
In summary, we developed MA-UITD, an offline multi-agent inverse reinforcement learning method that uses belief-conditioned successor features to recover explicit agent-specific rewards from fixed demonstrations in continuous-state environments with discrete or bounded continuous actions. The current method conditions reward inference on predictive beliefs estimated from observed behavior; an important direction for future work is to infer each agent’s internal beliefs jointly with its reward function and to extend the framework to interactions involving larger numbers of agents. One limitation is that demonstrations provide little evidence about rewards outside the support of the observed states. Reward estimates in these regions therefore depend largely on regularization and the inductive biases of the chosen function approximator, and should be interpreted cautiously until supported by additional data.

\subsection*{AI use statement}
In this work, we used generative AI tools to assist with language polishing, literature retrieval and discovery (e.g., identifying related work), and coding. It is important to note that generative AI tools were not involved in the ideation, research methodology, or experimental design. We have reviewed all AI-assisted work. We take responsibility for the final content of this work, including text, claims or artifacts produced with the aid of generative AI.

\bibliography{references}
\bibliographystyle{iclr2027_conference}

\appendix

\section{Algorithms}
\begin{algorithm}[H]
\caption{Multi-Agent Unilateral Inverse Temporal Difference Learning (MA-UITD)}
\label{alg:ma-uitd}
\begin{algorithmic}[1]
\Statex \textbf{Input:} Reward-free joint demonstrations $\mathcal D$;
number of agents $n$; discount $\gamma$; negatives $K$; temperature
$\tau_{\mathrm{NCE}}$; proposal
distributions $\{\nu_i\}_{i=1}^n$; learning rate $\alpha$; regularization
coefficient $\lambda_w$.
\Statex \textbf{Output:} Shared reward feature parameters
$\boldsymbol\theta^\Phi$ and agent-specific parameters
$\{\boldsymbol\eta_i,w_i\}_{i=1}^{n}$.

\State Initialize $\boldsymbol\theta^\Phi$,
$\{\boldsymbol\eta_i,w_i\}_{i=1}^{n}$, and target parameters
$\{\tilde{\boldsymbol\eta}_i\}_{i=1}^{n}$
\While{training budget remains}
    \State Sample a minibatch $\mathcal B\sim\mathcal D$
    \For{$i=1,\ldots,n$}
        \State Sample $K$ unilateral alternatives from
        $\nu_i(\cdot\mid s)$
        \State Compute $\mathcal L_i^{\mathrm{NCE}}$ using
        Equation~\eqref{eq:bc-nce-loss}
        \State $\boldsymbol\eta_i
        \gets\boldsymbol\eta_i
        -\alpha\nabla_{\boldsymbol\eta_i}
        \mathcal L_i^{\mathrm{NCE}}$
        \State $w_i\gets w_i-\alpha\nabla_{w_i}
        \bigl(\mathcal L_i^{\mathrm{NCE}}
        +\lambda_w\lVert w_i\rVert_1\bigr)$
    \EndFor
    \State Compute $\mathcal L^{\mathrm{UITD}}
    =n^{-1}\sum_{i=1}^{n}\mathcal L_i^{\mathrm{UITD}}$ using
    Equation~\eqref{eq:UITD_loss}
    \State $\boldsymbol\theta^\Phi\gets\boldsymbol\theta^\Phi
    -\alpha\nabla_{\boldsymbol\theta^\Phi}
    \mathcal L^{\mathrm{UITD}}$
    \For{$i=1,\ldots,n$}
        \State $\boldsymbol\eta_i
        \gets\boldsymbol\eta_i
        -\alpha\nabla_{\boldsymbol\eta_i}
        \mathcal L^{\mathrm{UITD}}$
        \State Update $\tilde{\boldsymbol\eta}_i$ from
        $\boldsymbol\eta_i$
    \EndFor
\EndWhile
\State \Return
$\widehat r_i(s)=\Phi(s;\boldsymbol\theta^\Phi)^\top w_i$ for all $i$
\end{algorithmic}
\end{algorithm}

\section{Implementation details}
For \textit{Simple Spread}, the reward-feature network takes the scalar agent--landmark and inter-agent distances as input, whereas the BCSF networks take the agent and landmark positions, agent velocities, and each agent's own action. For \textit{Simple Push}, the reward-feature network takes the agent and landmark positions as input, while the BCSF networks additionally receive the agent velocities and each agent's own action. For \textit{Gems}, a shared torso network, MLP$([32,32])$, encodes the agent positions. The reward-feature head operates on this shared state representation, while each BCSF head additionally receives the corresponding agent's own action. To ensure a fair comparison, all algorithms were provided with the same task-specific state and action information.

\subsection{Expert demonstrations.} For the discrete-action versions of \textit{Simple Spread} and \textit{Simple Push}, we generate demonstrations using expert policies trained with the multi-agent ACKTR implementation of \citet{yu_multi-agent_2019}. For their continuous-action versions and for both versions of the \textit{Gems} game, we use expert policies trained with MASAC \citep{bettini2024benchmarl}. The \textit{Gems} datasets contain an equal mixture of fully cooperative and fully competitive expert behavior. Across tasks, the demonstration datasets contain between 500 and 4,000 episodes.

\subsection{MA-UITD on discrete action datasets}
MA-UITD naturally generalizes to datasets with discrete action spaces by simply replacing NCE-loss in Eq.~\eqref{eq:bc-nce-loss} with a Maximum Likelihood Estimation(MLE) loss
\begin{equation}
\mathcal{L}^{\mathrm{MLE}}_i(\boldsymbol\eta_i,w_i)
=
-\mathbb{E}_{(s,a_i)\sim\mathcal D}
\left[
\log
\frac{
\exp\!\left(\bar{\Psi}_i(s,a_i;\boldsymbol\eta_i)^\top w_i/\tau_{\mathrm{MLE}}\right)
}{
\sum_{\tilde a_i\in\mathcal A_i}
\exp\!\left(\bar{\Psi}_i(s,\tilde a_i;\boldsymbol\eta_i)^\top
w_i/\tau_{\mathrm{MLE}}\right)
}
\right].
\end{equation}

\subsection{Noise distribution}
Let $\rho_E(s,\mathbf a)$ denote the state--joint-action occupancy induced
by the expert demonstrations, and let
\begin{equation}
\rho_{E,i}(s,a_i)
=
\int \rho_E(s,\mathbf a)\,\mathrm d\mathbf a_{-i}
\end{equation}

be its marginal for agent $i$. We draw a transition
$(s',\mathbf a')$ uniformly from $\mathcal D_{\mathrm{train}}$ and use
$a_i'$ as a negative action at the current state $s$. Equivalently, the
noise distribution is the expert's marginal action distribution,
\begin{equation}
\zeta_i(a_i)
=
\int \rho_{E,i}(s',a_i)\,\mathrm ds',    
\end{equation}

approximated empirically using the training demonstrations.

\subsection{Continuous policy rollout}
Rollout initial states were generated using a
fixed task-specific random seed, independently of training, and were
shared across all model seeds. At each state, we approximated each
agent's continuous Boltzmann policy using the same candidate-based
procedure. Specifically, we generated $K=256$ low-discrepancy candidate
actions from the agent's bounded action domain and evaluated
\begin{equation}
        \widehat{Q}_i(s,a_i)
    =
    \overline{\Psi}_i(s,a_i)^\top w_i .
\end{equation}

The executed action was sampled from the categorical approximation
\begin{equation}
        \widehat{\pi}_i(a_i\mid s)
    =
    \frac{\exp\!\left(\widehat{Q}_i(s,a_i)/\tau\right)}
         {\sum_{a'_i\in\mathcal{A}_{i,K}}
          \exp\!\left(\widehat{Q}_i(s,a'_i)/\tau\right)},
    \qquad \tau=1 ,
\end{equation}

where $\mathcal{A}_{i,K}$ denotes the candidate set. Candidate-generation
and policy-sampling seeds were fixed across model seeds, yielding paired
evaluation conditions. 
\subsection{Hyperparameters}
We conducted a comprehensive hyperparameter search for all three algorithms included in the comparison. Hyperparameters used by MA-UITD implementations are listed in the table~\ref{tab:ma-uitd-discrete-hyperparameters}, ~\ref{tab:ma-uitd-continuous-hyperparameters}, ~\ref{tab:ma-uitd-wz-hyperparameters}.
\begin{table}[H]
\centering
\caption{MA-UITD hyperparameters for discrete action datasets}
\label{tab:ma-uitd-discrete-hyperparameters}
\scriptsize
\setlength{\tabcolsep}{2pt}
\renewcommand{\arraystretch}{1.10}

\begin{tabularx}{\columnwidth}{@{}p{0.24\columnwidth}XXX@{}}

  \toprule
  Hyperparameter
    & Simple Spread
    & Simple Push
    & Gems \\
  \midrule

  Reward-feature approximator $\Phi$
    & MLP$([128, 128])$
    & MLP$([128,128])$
    & MLP$([64,32])$ \\

  BCSF approximator $\bar{\Psi}_i$
    & MLP$([256,256])$
    & MLP$([256,256])$
    & MLP$([64,64])$ \\

  Activation function
    & SiLU
    & SiLU
    & LeakyReLU, slope $0.01$ \\

  $\ell_1$ coefficient
    & $0.05$
    & $0.05$
    & $0.001$ \\

  Dropout rate
    & $0$
    & $0$
    & $0.3$ \\

  \midrule

  Number of dimensions in $\Phi$
    & $6$
    & $128$
    & $10$ \\

  \midrule

  $n$-step return
    & $4$
    & $4$
    & $1$ \\

  Discount factor $\gamma$
    & $0.95$
    & $0.95$
    & $0.99$ \\

  Polyak coefficient $\tau$
    & $10^{-4}$
    & $10^{-4}$
    & $10^{-4}$ \\
    
  Rationality $\tau_{\mathrm{MLE}}$
    & $1$
    & $1$
    & $1$ \\
  Optimiser
    & AdamW\citep{loshchilov_adamw_2019} for $\Phi,\bar{\Psi}_i$ (lr$=10^{-4}$, wd$=10^{-5}$); Adam\citep{kingma_adam_2017} for $w_i$(lr$=10^{-4}$)
    & AdamW\citep{loshchilov_adamw_2019} for $\Phi,\bar{\Psi}_i$ (lr$=10^{-4}$, wd$=10^{-5}$); Adam\citep{kingma_adam_2017} for $w_i$(lr$=10^{-4}$)
    & AdamW\citep{loshchilov_adamw_2019} for $\Phi,\bar{\Psi}_i$ and Adam\citep{kingma_adam_2017} for $w_i$; BC lr$=10^{-4}$, UITD lr$=5\!\times\!10^{-4}$ \\

  \bottomrule
\end{tabularx}
\end{table}

\begin{table}[H]
\centering
\caption{MA-UITD hyperparameters for continuous-action datasets}
\label{tab:ma-uitd-continuous-hyperparameters}
\scriptsize
\setlength{\tabcolsep}{2pt}
\renewcommand{\arraystretch}{1.10}

\begin{tabularx}{\columnwidth}{@{}p{0.24\columnwidth}XXX@{}}

  \toprule
  Hyperparameter
    & Simple Spread
    & Simple Push
    & Gems \\
  \midrule

  Reward-feature approximator $\Phi$
    & MLP$([128,128])$
    & MLP$([128,128])$
    & MLP$([64,32])$ \\

  BCSF approximator $\bar{\Psi}_i$
    & MLP$([256,256])$
    & MLP$([256,256])$
    & MLP$([128,64])$ \\

  Activation function
    & SiLU
    & SiLU
    & LeakyReLU, slope $0.02$ \\

  $\ell_1$ coefficient
    & $0.05$
    & $0.05$
    & $0.001$ \\

  Dropout rate
    & $0$
    & $0$
    & $0.3$ \\

  \midrule

  Number of dimensions in $\Phi$
    & $64$
    & $128$
    & $16$ \\

  \midrule

  $n$-step return
    & $4$
    & $4$
    & $1$ \\

  Discount factor $\gamma$
    & $0.95$
    & $0.95$
    & $0.99$ \\

  Polyak coefficient $\tau$
    & $10^{-4}$
    & $10^{-4}$
    & $10^{-4}$ \\

  Rationality $\tau_{\mathrm{NCE}}$
    & $1$
    & $1$
    & $1$ \\
    
  Optimiser
    & AdamW\citep{loshchilov_adamw_2019} for $\Phi,\bar{\Psi}_i$ (lr$=10^{-4}$, wd$=10^{-5}$); Adam\citep{kingma_adam_2017} for $w_i$(lr$=10^{-4}$)
    & AdamW\citep{loshchilov_adamw_2019} for $\Phi,\bar{\Psi}_i$ (lr$=10^{-4}$, wd$=10^{-5}$); Adam\citep{kingma_adam_2017} for $w_i$(lr$=10^{-4}$)
    & AdamW\citep{loshchilov_adamw_2019} for $\Phi,\bar{\Psi}_i$ and Adam\citep{kingma_adam_2017} for $w_i$; BC lr$=10^{-4}$, UITD lr$=5\!\times\!10^{-4}$ \\

  \bottomrule
\end{tabularx}
\end{table}

\begin{table}[H]
\centering
\caption{MA-UITD hyperparameters for real animal trajectories}
\label{tab:ma-uitd-wz-hyperparameters}
\scriptsize
\setlength{\tabcolsep}{3pt}
\renewcommand{\arraystretch}{1.10}

\begin{tabularx}{\columnwidth}{@{}p{0.34\columnwidth}X@{}}

  \toprule
  Hyperparameter &  \\
  \midrule

  State representation
    & Physical-order neck positions $(x_1,y_1,x_2,y_2)$ \\

  Action representation
    & Own observed 2-D neck displacement \\

  Reward-feature approximator $\Phi$
    & Position MLP$([128,64])$; distance MLP$([64,32])$ \\

  BCSF approximator $\bar{\Psi}_i$
    & MLP$([256,128])$\\

  Activation function
    & LeakyReLU, slope $0.01$ \\

  $\ell_1$ coefficient
    & $0.05$ \\

  Dropout rate
    & $0.3$ for both $\Phi$ and $\bar{\Psi}_i$ \\

  \midrule

  Number of dimensions in $\Phi$
    & $20$ ($8+8+4$) \\

  \midrule

  $n$-step return
    & $1$ \\

  Discount factor $\gamma$
    & $0.99$ \\

  Polyak coefficient $\tau$
    & $10^{-4}$ \\

  Optimiser
    & AdamW\citep{loshchilov_adamw_2019} for $\Phi,\bar{\Psi}_i$ (lr$=10^{-4}$, wd$=10^{-5}$); Adam\citep{kingma_adam_2017} for $w_i$(lr$=10^{-4}$) \\

  \bottomrule
\end{tabularx}
\end{table}

\subsection{Compute Resources}
All experiments were conducted on a workstation equipped with a 12-core AMD Ryzen~9 9900X CPU, 64\,GB of system memory, and a single NVIDIA GeForce RTX~5080 GPU with 16\,GB of memory. Each training run used one GPU. Experiments were implemented in Python~3.10.20 using PyTorch~2.11.0 and CUDA~12.8.
\section{Extended Results}
\subsection{Reward recovery for MA-UITD and baselines}
Details for reward recovery performance across MA-UITD and baselines are shown in Table~\ref{tab:PCC/SCC_comparison}
\begin{table}[htbp]
  \centering
  \caption{Held-out immediate-reward recovery.}
  \label{tab:PCC/SCC_comparison}
  \small
  \setlength{\tabcolsep}{7pt}
  \begin{tabular}{@{}lcccc@{}}
    \toprule
    Method
      & Simple Spread
      & \multicolumn{2}{c}{Simple Push}
      & Gems \\
    \cmidrule(lr){3-4}
      & \shortstack{PCC $\uparrow$ \\ SCC $\uparrow$}
      & \shortstack{Adversary PCC $\uparrow$\\ Adversary SCC$\uparrow$}
      & \shortstack{Good PCC $\uparrow$\\ Good SCC$\uparrow$}
      & \shortstack{PCC $\uparrow$ \\ SCC $\uparrow$} \\
    \midrule
    MA-AIRL
      & \shortstack{$0.641 \pm 0.014$ \\ $0.745 \pm 0.013$}
      & \shortstack{$0.178 \pm 0.022$ \\ $0.145 \pm 0.024$}
      & \shortstack{$0.581 \pm 0.057$ \\ $0.375 \pm 0.070$}
      & \shortstack{$0.230 \pm 0.021$ \\ $0.145 \pm 0.021$} \\
    MAMQL
      & \shortstack{$-0.140 \pm 0.016$ \\ $-0.154 \pm 0.019$}
      & \shortstack{$0.081 \pm 0.012$ \\ $-0.000 \pm 0.019$}
      & \shortstack{$-0.003 \pm 0.003$ \\ $-0.074 \pm 0.014$}
      & \shortstack{$0.142 \pm 0.029$ \\ $0.174 \pm 0.037$} \\
    MA-UITD
      & \shortstack{$\mathbf{0.940 \pm 0.008}$ \\ $\mathbf{0.942 \pm 0.007}$}
      & \shortstack{$\mathbf{0.564 \pm 0.184}$ \\ $\mathbf{0.372 \pm 0.138}$}
      & \shortstack{$\mathbf{0.854 \pm 0.016}$ \\ $\mathbf{0.554 \pm 0.040}$}
      & \shortstack{$\mathbf{0.711 \pm 0.014}$ \\ $\mathbf{0.321 \pm 0.009}$} \\
    \bottomrule
  \end{tabular}
  \vspace{2pt}

  {\footnotesize Entries are mean $\pm$ sample standard deviation across three training seeds on the fixed 500-episode test split; the upper and lower values in each cell are PCC and SCC, respectively. }
\end{table}

\subsection{Policy imitation for the \textit{Gems}}
Details for policy imitation in \textit{Gems} are shown in Table~\ref{tab:Gems_collection_distribution}
\begin{table}[htbp]
    \centering
    \small
    \setlength{\tabcolsep}{8pt}
    \caption{Policy imitation for the \textit{Gems}}
    \begin{tabular}{lcc}
        \toprule
        Method & Purple gems (\%) & Red/Blue gems (\%) \\
        \midrule
        \textbf{Ground Truth} & 50 & 50 \\
        MA-AIRL          & $0.0 \pm 0.0$   & $100.0 \pm 0.0$ \\
        MAMQL            & $50.9 \pm 10.7$ & $49.1 \pm 10.7$ \\
        \textbf{MA-UITD} & $44.8 \pm 1.4$  & $55.2 \pm 1.4$ \\
        \bottomrule
    \end{tabular}
    \label{tab:Gems_collection_distribution}
    \\
    {\footnotesize Percentage of gems collected. Values are reported as the mean $\pm$ sample standard deviation across three training seeds, each evaluated over 500 stochastic-policy episodes.}
\end{table}
\subsection{Reward visualization}
\begin{figure}[H]
    \centering
    \includegraphics[
        width=\linewidth,]{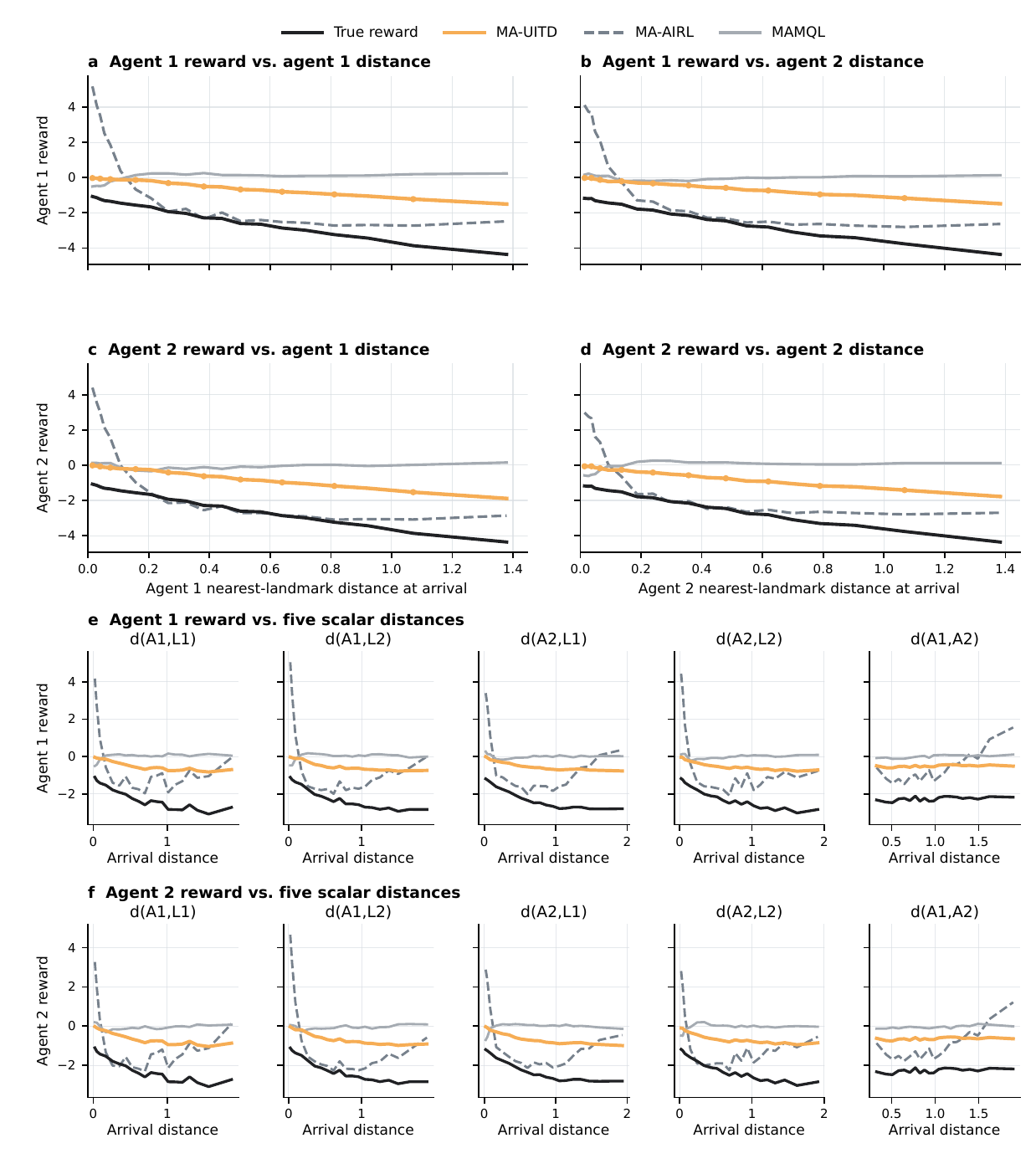}
    \caption{\textbf{Reward recovery in discrete \textit{Simple Spread}.}
\textbf{a,b}, Agent $1$ reward versus each agent’s nearest-landmark distance; \textbf{c,d}, corresponding Agent $2$ results.
\textbf{e,f}, Agent $1$ and Agent $2$ rewards versus the five indicated arrival-state distances.
Curves show the true and raw MA-UITD, MA-AIRL, and MAMQL rewards, averaged in equal-width bins over 4,836 held-out transitions (500 episodes).}
\label{fig:simplre_spread_visual_discrete}
\end{figure}

\begin{figure}[H]
    \centering
    \includegraphics[
        width=\linewidth,]{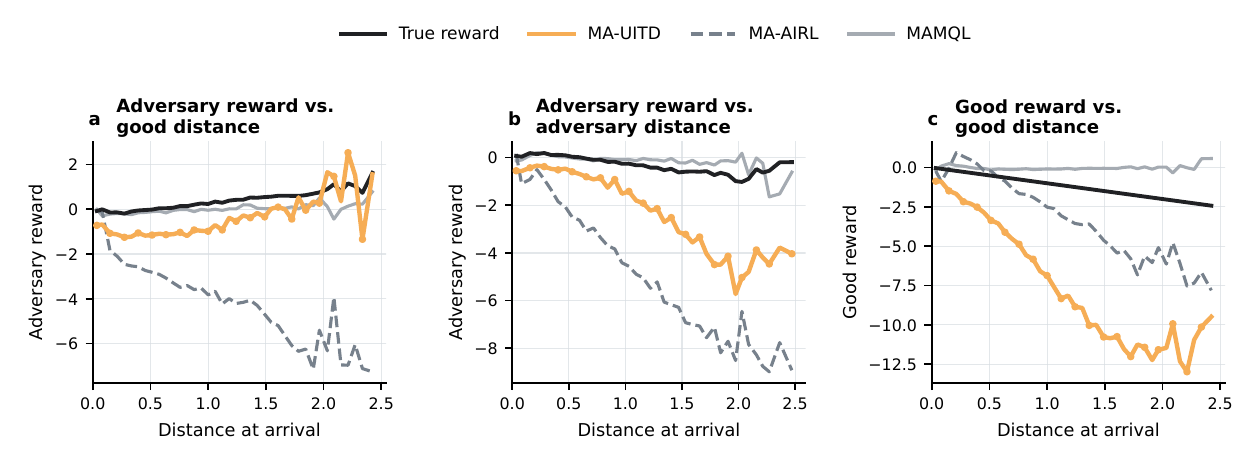}
    \caption{\textbf{Reward recovery in discrete \textit{Simple Push}.}
\textbf{a,b}, Adversary reward versus the good agent’s and its own landmark distances, respectively.
\textbf{c}, Good-agent reward versus its landmark distance.
Curves show the true and raw MA-UITD, MA-AIRL, and MAMQL rewards, averaged in equal-width bins over 12,500 held-out transitions (500 episodes).}
\label{fig:simple_push_visual_discrete}
\end{figure}

\begin{figure}[H]
    \centering
    \includegraphics[
        width=\linewidth,]{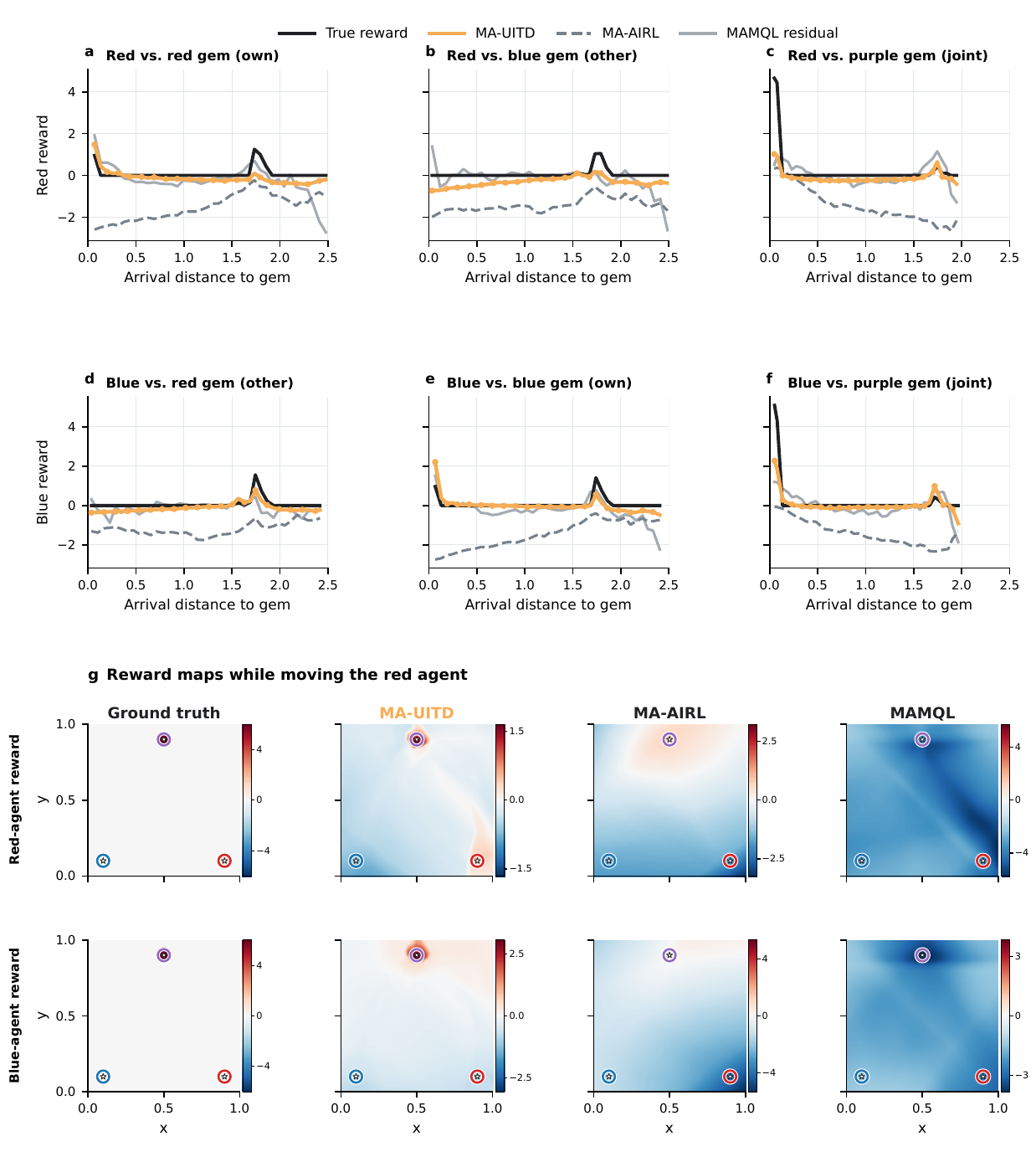}
    \caption{\textbf{Reward recovery in discrete \textit{Gems}.}
\textbf{a--c}, Red-agent reward versus arrival distance to the red, blue, and purple gems; \textbf{d--f}, corresponding blue-agent results. Curves compare ground truth with raw MA-UITD, MA-AIRL, and MAMQL rewards, averaged in equal-width bins over 10,274 held-out transitions (500 episodes). \textbf{g}, Spatial reward maps obtained by moving the red agent while fixing the blue agent at the purple gem. Rows show red- and blue-agent rewards; columns show ground truth and the three recovered rewards. Circles mark gem locations and collection radii. Each map uses an independent symmetric scale centered at zero (blue, negative; red, positive).}
\label{fig:Gems_reward_visual_discrete}
\end{figure}

\begin{figure}[H]
    \centering
    \includegraphics[
        width=\linewidth,]{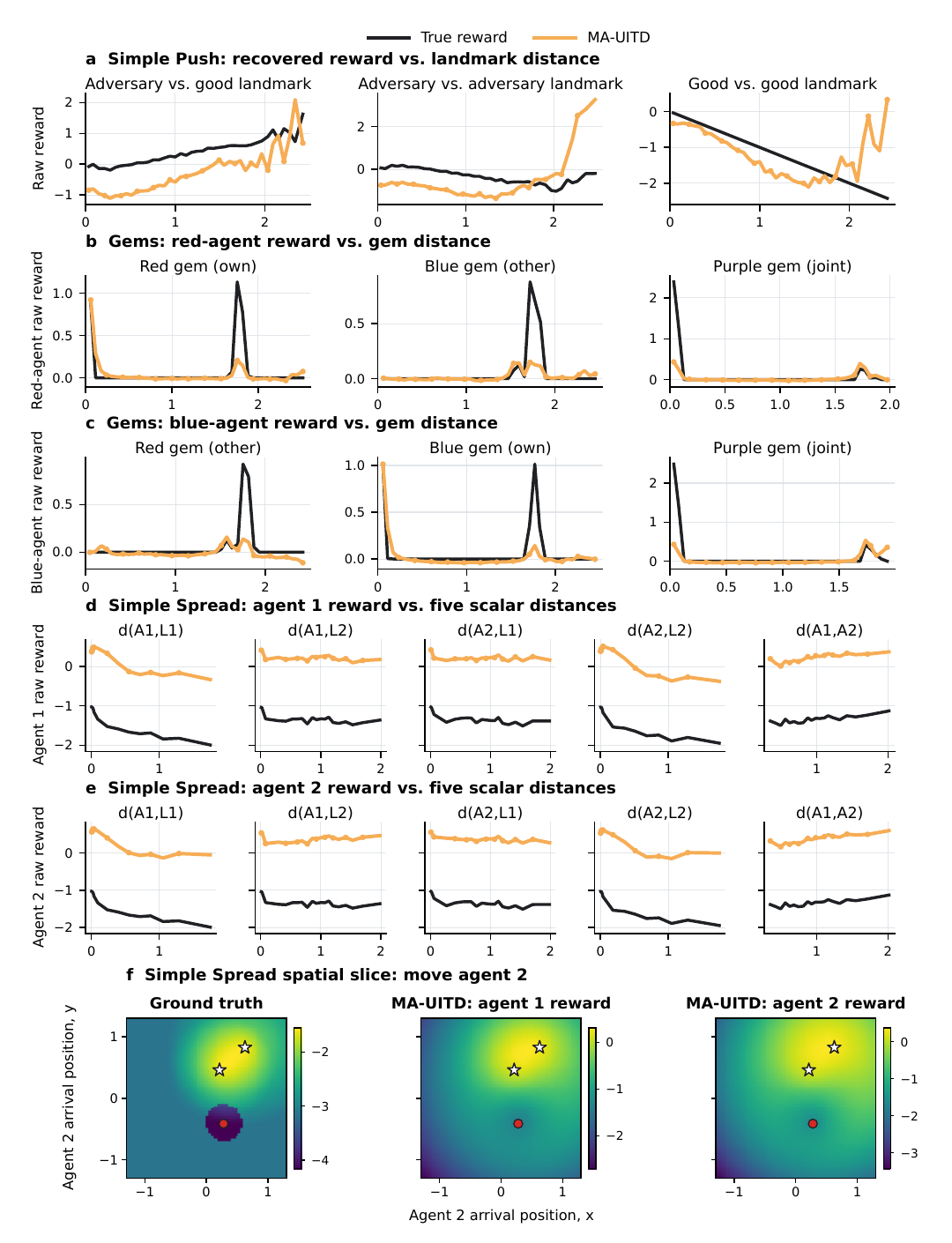}
    \caption{\textbf{Reward recovery across continuous-action environments.}
\textbf{a}, Reward--distance relationships in \textit{Simple Push}.
\textbf{b,c}, Red- and blue-agent relationships in \textit{Gems}.
\textbf{d,e}, Agent $1$ and Agent $2$ rewards versus five arrival-state distances in \textit{Simple Spread}.
\textbf{f}, Spatial rewards obtained by moving Agent $2$ while fixing Agent $1$ and the landmarks.
Black and orange denote true and raw MA-UITD rewards. Curves are binned means from 500 held-out episodes per task; Spatial maps use independent raw color scales.}
\label{fig:visual_ctn}
\end{figure}

In this section, we provide a comprehensive visual analysis of the reward functions recovered from both discrete- and continuous-action datasets(Fig.~\ref{fig:simplre_spread_visual_discrete}, ~\ref{fig:simple_push_visual_discrete}, ~\ref{fig:Gems_reward_visual_discrete}, ~\ref{fig:visual_ctn}).

\subsubsection{Recovered reward in discrete-action environments}

\label{appendix_discrete_reward_visualizations}

\subsubsection{Recovered reward in continuous-action environments}

\label{appendix_continuous_reward_visualizations}

\end{document}